\documentclass{article} 
\usepackage{iclr2027_conference,times}
\iclrfinalcopy

\usepackage{amsmath,amsfonts,bm}

\def\eqref#1{equation~\ref{#1}}

\def\1{\bm{1}}

\DeclareMathAlphabet{\mathsfit}{\encodingdefault}{\sfdefault}{m}{sl}
\SetMathAlphabet{\mathsfit}{bold}{\encodingdefault}{\sfdefault}{bx}{n}

\usepackage{graphicx}
\usepackage{hyperref}
\usepackage{booktabs} 
\usepackage{multirow} 
\usepackage{url}
\usepackage{subcaption}

\title{Reliable Parallel Decoding in Masked\\ Diffusion Language Models}

\author{%
  \textbf{Zhenghao He}\textsuperscript{$\dagger$}\quad
  \textbf{Bohan Liu}\\
  \textbf{Guangzhi Xiong}\quad
  \textbf{Aidong Zhang}\textsuperscript{$\dagger$}\\[5pt]
  Department of Computer Science, University of Virginia\\
  \textsuperscript{$\dagger$}\texttt{\{zhenghao, aidong\}@virginia.edu}
}

\usepackage{xcolor}
\usepackage[table]{xcolor}
\usepackage{enumitem}

\begin{document}

\maketitle

\begin{abstract}
Masked diffusion language models (MDLMs) can generate text efficiently by predicting multiple masked tokens in parallel, but predictions from the same forward pass are not necessarily reliable when committed together. 
We study when parallel commitment is reliable. 
Our diagnostics show that confidence alone does not determine a reliable commitment order: confident predictions near the end of the sequence can fix an answer before its supporting computations are established, and downstream predictions become less reliable as the uncertainty of their upstream context grows. At the same time, a single forward pass can already resolve several masked tokens, and predictions that remain stable across the final layers are more likely to be correct. Based on these findings, we propose Reliable Parallel Decoding (RPD), a training-free method that selects candidates by layerwise prediction stability and final confidence, and commits them under a cumulative entropy budget over their preceding masked positions. RPD defers predictions with uncertain upstream context while committing the remaining candidates in parallel, without relying on a fixed block schedule. Across mathematical reasoning and code generation benchmarks on LLaDA and Dream, RPD achieves the highest decoding throughput among the evaluated methods while maintaining or improving accuracy.
\end{abstract}

\section{Introduction}

Masked diffusion language models (MDLMs) offer the potential for efficient generation by predicting multiple masked tokens in parallel and progressively filling a sequence over successive decoding iterations~\citep{nie2026large, gong2022diffuseq}. At each iteration, the decoder must choose which predictions to accept and use as context for the next iteration. Committing more tokens can reduce the number of decoding iterations, but predictions produced in the same forward pass are not necessarily reliable when committed together. 
These considerations motivate our central research question: \textbf{\emph{When can masked diffusion language models safely commit multiple tokens in parallel during decoding?}}

\begin{figure}[t]
    \centering
    \includegraphics[width=\linewidth]{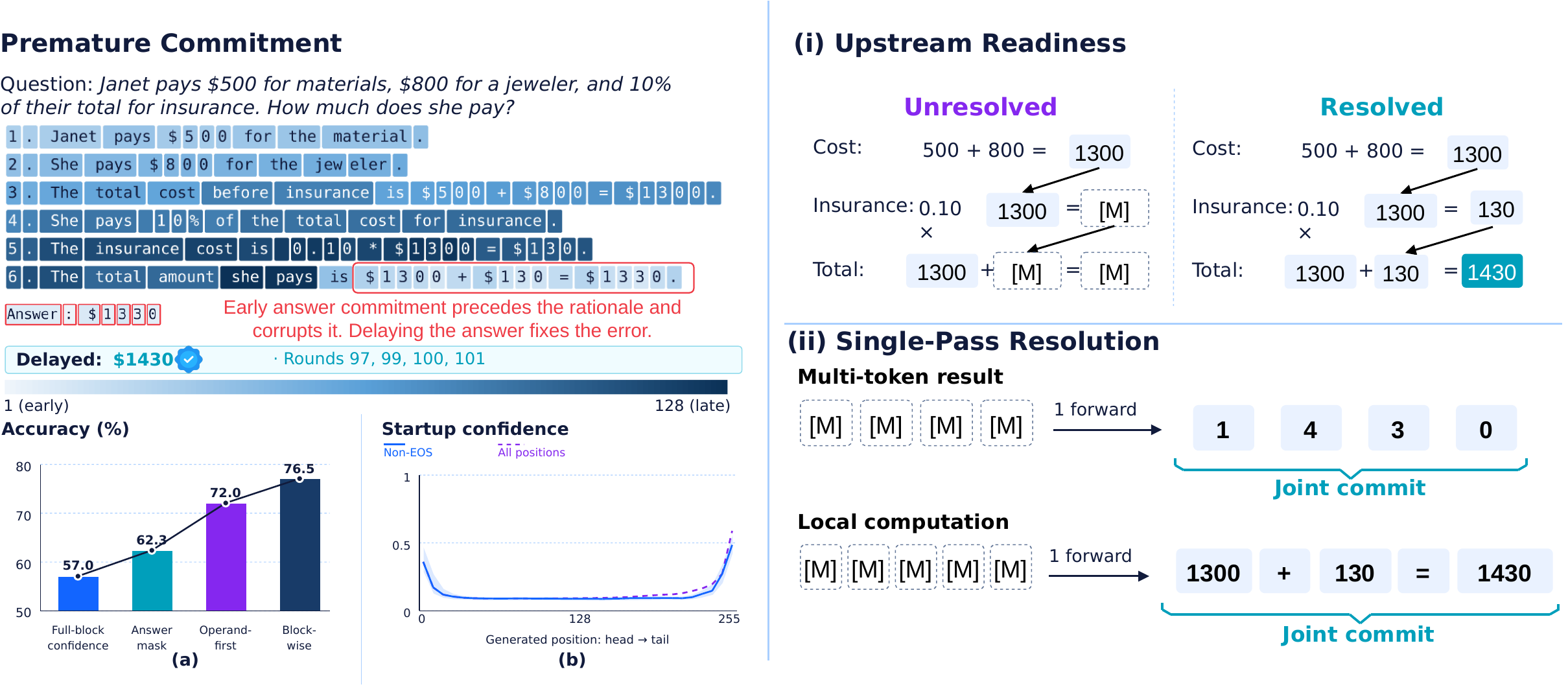}
    \caption{
Left: Confidence-based decoding commits the answer early and fixes an incorrect value (\$1330), whereas delaying the answer yields the correct \$1430. (a) Accuracy improves under stricter ordering constraints. (b) Initial confidence peaks near the sequence boundaries. Right: (i) Downstream tokens should wait until their upstream values are resolved. (ii) Tokens resolved within one forward pass can be committed together.
}
\vspace{-5pt}
    \label{fig:intro}
\end{figure}

Figure~\ref{fig:intro} Left (b) shows a U-shaped confidence profile across the generation canvas, with high confidence near both ends of the sequence. High confidence at the end reflects strong formatting priors rather than a correctly computed answer. As a result, confidence-based decoding can lock in an incorrect answer before its supporting computations are established (Figure~\ref{fig:intro} Left). Imposing ordering constraints mitigates this problem (Figure~\ref{fig:intro} Left (a)): delaying the answer region until all other tokens are committed improves accuracy from 57.0\% to 62.3\%, and further requiring operands to precede their dependent results raises it to 72.0\%. These results suggest that confidence alone does not determine a reliable commitment order. A reliable order should instead reflect dependencies: a prediction should wait while the information it depends on remains unresolved, and can otherwise be committed in parallel. Existing methods approximate this in different ways.

Block-wise decoding, as used in LLaDA~\citep{nie2026large}, imposes order at a fixed granularity, decoding in parallel within each block while enforcing order across blocks, but its boundaries do not adapt to which dependencies remain unresolved. Other methods select tokens adaptively but judge each prediction without regard to its context. Fast-dLLM~\citep{wu2026fast} and EB-Sampler~\citep{ben2026accelerated} use confidence thresholds or entropy budgets over the selected tokens, without assessing whether the upstream information supporting each prediction has been resolved. DAPD~\citep{kim2026dapd} uses attention to avoid committing strongly coupled tokens together, but separating dependent predictions does not determine which should come first. LoPA~\citep{xu2025lopascalingdllminference} explores alternative filling orders to improve subsequent parallelism, at the cost of evaluating additional branches. What remains missing is a way to use the current forward pass to decide which predictions should wait and which can be committed together.


We address this question with two complementary insights. First, downstream predictions should wait while their upstream information remains uncertain. In Figure~\ref{fig:intro} (i), committing the total while the insurance amount is still masked risks locking in an unsupported answer. Second, a single forward pass can already resolve several masked tokens at once. In Figure~\ref{fig:intro} (ii), one forward pass produces all digits of $1430$ and even the local computation $1300+130=1430$. Section~\ref{sec:diagnosis} shows that predictions remaining consistent across the final layers without losing probability are more likely to be correct, which allows such tokens to be identified and committed together.

We turn these insights into Reliable Parallel Decoding (RPD), a method that uses upstream uncertainty and prediction stability to decide which tokens to commit at each decoding step. RPD delays predictions whose upstream information remains too uncertain and uses layerwise consistency together with final confidence to identify candidates for joint commitment. Across four mathematical reasoning and code generation benchmarks on LLaDA and Dream, RPD achieves the highest decoding throughput in all eight model--task settings, running 2.4--6.1$\times$ faster than default decoding, while maintaining or improving accuracy.

Our main contributions are:
\begin{itemize}
    \item We analyze parallel commitment in MDLMs and identify two properties that determine its reliability: downstream predictions are less reliable when their upstream context remains uncertain, and predictions that remain stable across layers are more likely to be correct.
    \item We propose Reliable Parallel Decoding (RPD), a training-free method that selects candidates by layerwise stability and commits them under a cumulative entropy budget, ordering commitment by upstream uncertainty rather than a fixed block schedule.
    \item We evaluate RPD on mathematical reasoning and code generation with LLaDA and Dream. RPD achieves the highest decoding throughput among the evaluated methods while maintaining or improving accuracy, and ablations show that layerwise stability and cumulative entropy contribute parallelism and reliability, respectively.
\end{itemize}

\begin{figure}[t]
    \centering
    \includegraphics[width=\linewidth]{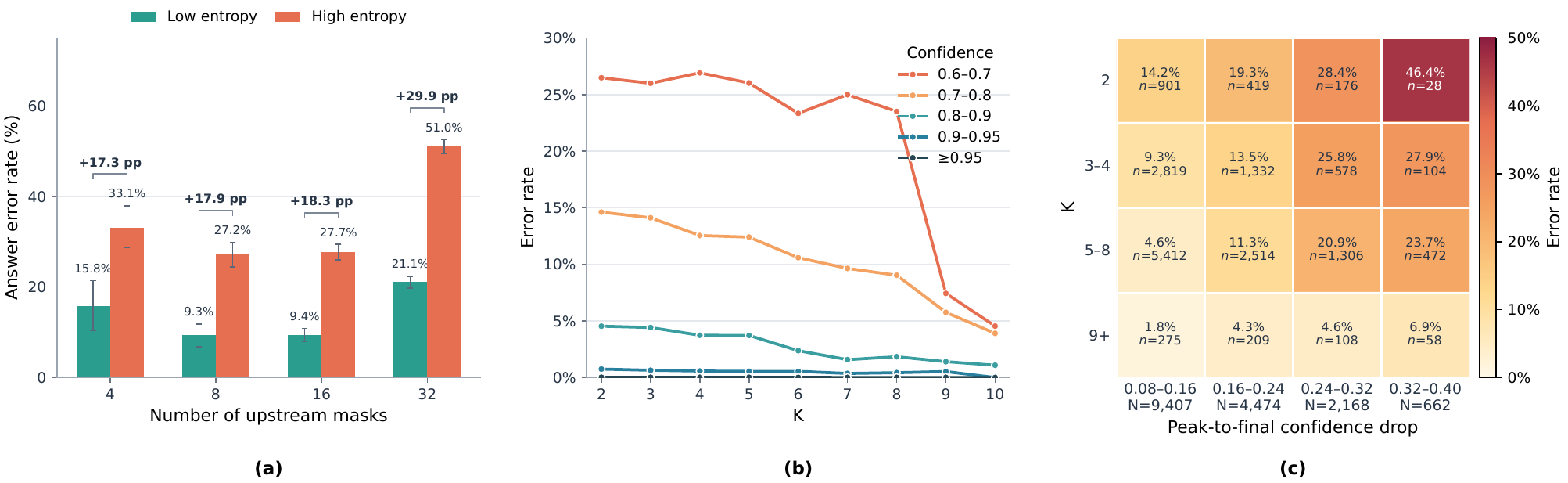}
\caption{
Diagnostics on Dream.
(a) Answer error rates for low- and high-upstream-entropy masking patterns within the same problem and mask count.
(b) Token error rates under increasing \emph{prediction persistence} $K$.
(c) Token error rates by  \emph{prediction persistence} $K$ and \emph{confidence drop} $r_i$ for predictions with final confidence in $[0.6,0.9)$.
}
\label{fig:diagnosis}
\end{figure}
\section{Understanding Parallel Commitment}
\label{sec:diagnosis}

The two insights above raise two questions: how unresolved upstream uncertainty relates to downstream errors, and how to identify predictions that can be reliably committed together.
We investigate these questions through controlled masking experiments and analysis of layerwise predictions.

We use 2,000 gold arithmetic traces from GSM8K-Aug~\citep{deng2024explicit}, masking both selected upstream digits and all digits of the final answer.
For each problem, we sample 16 masking patterns at each upstream mask count $m\in\{4,8,16,32\}$ and evaluate each state with one forward pass.
Data construction and complete results, including LLaDA, are given in~\ref{app:diagnostic_setup} and~\ref{app:diagnostic_results}

\subsection{Upstream Uncertainty and Downstream Errors}
\label{sec:upstream_uncertainty}

Figure~\ref{fig:intro} Left (b) shows how high-confidence predictions near the end of the sequence can be committed before their supporting calculations are complete.
LLaDA's block-wise decoding prevents commitment in later blocks, while Dream's default configuration, with temperature $0.1$, top-$p$ $0.9$, and lowest-entropy-first selection, produces an approximately left-to-right order in our observations (Appendix~\ref{app:commitment_order}).
Both give earlier computations time to be established before downstream predictions are committed.
This raises a question: \emph{are downstream errors associated with uncertainty in the upstream information that remains unresolved?}

We define \emph{cumulative entropy} as
\begin{equation}
E=\sum_{j\in\mathcal{U}}H(p_j),
\qquad
H(p_j)=-\sum_{v\in\mathcal{V}}p_j(v)\log p_j(v).
\label{eq:cumulative_entropy}
\end{equation}
where $\mathcal{U}$ contains the masked upstream digit positions, $p_j$ is the predictive distribution at position $j$, and $\mathcal{V}$ is the vocabulary.
Within each problem and mask count, we compare the five lowest-entropy masking patterns with the five highest-entropy patterns, retaining problems with both correct and incorrect answers.
Figure~\ref{fig:diagnosis}a shows that high-entropy patterns have answer error rates 17.3--29.9 percentage points higher, with all paired 95\% confidence intervals above zero.
These results support our first insight: downstream predictions are less reliable when their upstream information remains uncertain.

\subsection{Layerwise Stability and Prediction Reliability}
\label{sec:layerwise_stability}

Respecting upstream dependencies does not mean that every token must wait for a separate decoding iteration.
Such a restriction would sacrifice the parallelism that distinguishes MDLMs from autoregressive models.
Our second insight is that a single forward pass can already resolve several masked tokens.
We therefore examine whether the evolution of predictions across layers helps identify these opportunities.

Using the same masked states, we project intermediate hidden states into the vocabulary space and analyze predictions at the target-answer positions over the final half of the network.
Let $y_i$ be the final-layer prediction.
We define \emph{prediction persistence} as
\begin{equation}
K_i=\max\left\{
k\in\{1,\ldots,L-\ell_0+1\}:
\arg\max_v p_i^{(\ell)}(v)=y_i
\;\; \text{for all } \ell=L-k+1,\ldots,L
\right\}.
\label{eq:prediction_persistence}
\end{equation}
where $p_i^{(\ell)}$ is the predictive distribution at layer $\ell$, $L$ is the final layer, and $\ell_0$ is the first layer in the analyzed half.
Thus, $K_i$ counts consecutive layers ending at the final layer that predict the same token.
Figure~\ref{fig:diagnosis}b plots token error rates under increasing persistence requirements $K_i\ge k$, grouped by final confidence.
Predictions that persist across more layers generally have lower error rates, even within the same confidence range.

Beyond how long a prediction persists, we also examine how its probability changes during this period.
Even when the predicted token stays the same, its probability can fall before the final layer.
We quantify this change as \emph{confidence drop},
\begin{equation}
r_i
=
\max_{\ell\in[s_i,L]}p_i^{(\ell)}(y_i)
-
p_i^{(L)}(y_i).
\label{eq:confidence_drop}
\end{equation}
where $p_i^{(\ell)}$ is the predictive distribution at layer $\ell$, $s_i$ starts the final consistent suffix, and $L$ is the final layer.
Figure~\ref{fig:diagnosis}c shows that larger confidence drops correspond to more errors, while longer persistence corresponds to fewer errors.
For $K_i=5$--$8$, the error rate rises from 4.6\% to 23.7\% as confidence drop increases. These findings guide the design of our Reliable Parallel Decoding (RPD) method in Section~\ref{sec:method}.

\section{Reliable Parallel Decoding}
\label{sec:method}
\begin{figure}[t]
    \centering
    \includegraphics[width=\linewidth]{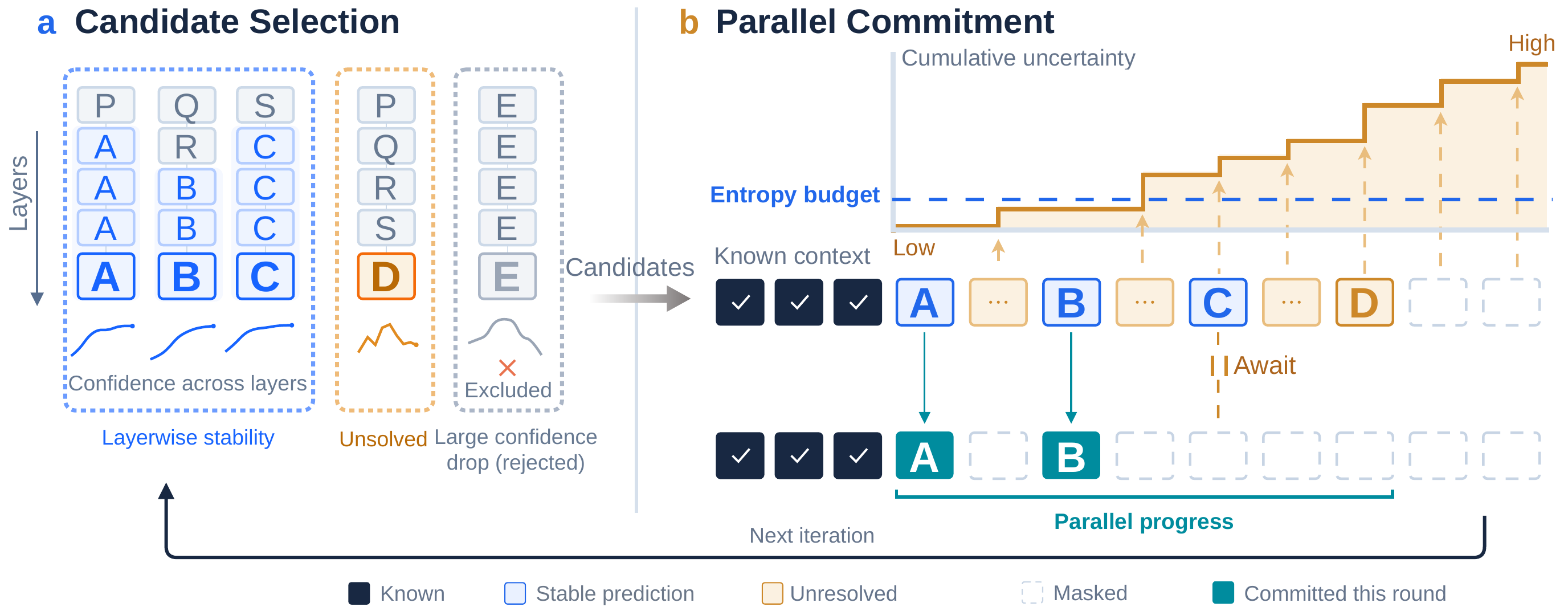}
    \caption{
    Overview of RPD.
    (a) Layerwise stability and confidence identify candidate predictions.
    Prediction persistence favors consistent predictions, while confidence drop penalizes predictions that lose probability.
    (b) A cumulative entropy budget determines which candidates can be committed together.
    $A$ and $B$ are committed in parallel, while $C$ waits because the cumulative entropy of its preceding masked positions exceeds the budget.
    }
    \label{fig:method}
\end{figure}

Section~\ref{sec:diagnosis} identifies two conditions for committing a prediction reliably: the prediction should be stable across layers (Section~\ref{sec:layerwise_stability}), and the upstream information it depends on should be sufficiently resolved (Section~\ref{sec:upstream_uncertainty}). Reliable Parallel Decoding (RPD) turns these conditions into two stages that operate on a single forward pass (Figure~\ref{fig:method}). Candidate selection (Section~\ref{sec:method_stability}) keeps predictions that are individually reliable, and parallel commitment (Section~\ref{sec:method_entropy}) determines which candidates can be committed together.

\subsection{Candidate Selection with Layerwise Stability}
\label{sec:method_stability}

Final confidence alone does not determine whether a prediction is reliable. Within the same confidence range, token error rates decrease as prediction persistence grows (Figure~\ref{fig:diagnosis}b) and increase with confidence drop (Figure~\ref{fig:diagnosis}c). We therefore combine the two layerwise quantities into a stability score:
\begin{equation}
S_i = \min(K_i, K_{\max}) - w\, r_i,
\label{eq:stability}
\end{equation}
where $K_{\max}$ caps the contribution of prediction persistence and $w > 0$ weights the penalty for confidence drop. Under a fixed threshold on $S_i$, longer persistence tolerates a larger drop, up to the cap. This follows Figure~\ref{fig:diagnosis}c: as confidence drop increases, the error rate rises from 14.2\% to 46.4\% for $K_i = 2$, but only from 1.8\% to 6.9\% for $K_i \geq 9$.

The stability test matters most at intermediate confidence. Predictions with very high confidence are already reliable (Figure~\ref{fig:diagnosis}b), whereas a low-confidence prediction can keep the same token across layers without being correct. Let $\mathcal{M}_t$ denote the masked positions at iteration $t$ and $c_i = p_i^{(L)}(y_i)$ the final confidence. We define the candidate set as
\begin{equation}
\mathcal{C}_t = \bigl\{\, i \in \mathcal{M}_t :
c_i \geq \theta_h
\ \text{or}\
\bigl(\theta_c \leq c_i < \theta_h \text{ and } S_i \geq \theta_s\bigr)
\,\bigr\},
\label{eq:candidate}
\end{equation}
where $\theta_h$ is the high-confidence threshold, $\theta_c < \theta_h$ is the minimum confidence for stability-based admission, and $\theta_s$ is the stability threshold. Hyperparameter values are given in Appendix~\ref{app:rpd_impl}. High-confidence predictions bypass only the stability test and remain subject to the cumulative entropy budget in Section~\ref{sec:method_entropy}. Figure~\ref{fig:method}a illustrates this selection. $A$, $B$, and $C$ remain consistent across the final layers and become candidates. $D$ keeps changing, and $E$ is excluded by a large confidence drop despite its persistent prediction.

\subsection{Parallel Commitment with Cumulative Entropy}
\label{sec:method_entropy}
Candidates in $\mathcal{C}_t$ have stable predictions, but stability alone does not ensure that a prediction is supported by resolved context. As shown in Figure~\ref{fig:intro} Left (b), a prediction near the end of the sequence can be confident and stable while the computations it depends on are still masked, and Section~\ref{sec:upstream_uncertainty} shows that such downstream predictions become less reliable as upstream uncertainty grows. We therefore commit a candidate only when the uncertainty in its preceding masked positions is sufficiently low.

Whether this condition holds depends not only on the current masks, but also on which earlier candidates are committed alongside it in the same iteration. We therefore construct the commitment set sequentially. Starting from an empty set, we scan the candidates in $\mathcal{C}_t$ from left to right. When the scan reaches candidate $i$, let $\mathcal{A}_{t,<i}$ denote the candidates already selected. We measure the upstream uncertainty of $i$ as the cumulative entropy of the preceding positions that will remain masked after this iteration:
\begin{equation}
E_i = \sum_{\substack{j<i \\ j \in \mathcal{M}_t \setminus \mathcal{A}_{t,<i}}} H\!\left(p_j^{(L)}\right),
\end{equation}
which mirrors Eq.~\ref{eq:cumulative_entropy} using the final-layer distributions of the current forward pass. The sum covers every preceding masked position, including non-candidates and rejected candidates, but excludes $\mathcal{A}_{t,<i}$: positions committed together with $i$ serve as its context, since related tokens can be resolved within a single forward pass (Section~\ref{sec:layerwise_stability}). Candidate $i$ is selected if $E_i \le \beta$ for a fixed entropy budget $\beta$, giving
\begin{equation}
\mathcal{A}_t = \{\, i \in \mathcal{C}_t : E_i \le \beta \,\}.
\end{equation}
If $\mathcal{A}_t$ is empty, we commit the single prediction with the highest final confidence among the first $W$ masked positions from the left, which guarantees progress at every iteration. All positions in $\mathcal{A}_t$ are then filled with their predictions $y_i$ from the same forward pass, and the remaining positions stay masked for the next iteration. In Figure~\ref{fig:method}b, $A$ and $B$ are committed together because little entropy separates them, while $C$ waits despite its stable prediction because the entropy accumulated before it exceeds $\beta$.

\paragraph{Variants.} Full RPD decodes over the full canvas without blocks, relying on the cumulative entropy budget to determine commitment order. We also consider RPD-block, which applies the candidate selection of Section~\ref{sec:method_stability} within fixed 32-token blocks and omits the cumulative entropy budget, relying on the block order alone. It isolates the contribution of layerwise stability and offers a faster operating point, while full RPD further delays candidates with uncertain upstream context.
\section{Experiments}
\subsection{Experimental Setup}
\label{sec:experimental-setup}

\paragraph{Datasets.}
We evaluate mathematical reasoning on the full test sets of GSM8K~\citep{cobbe2021trainingverifierssolvemath} and MATH-500~\citep{hendrycks2021measuring,lightman2024let}, and code generation on
HumanEval~\citep{chen2021evaluatinglargelanguagemodels} and MBPP~\citep{austin2021programsynthesislargelanguage}.  We use fixed, task-specific zero-shot
prompts, the models' native chat templates, and a maximum generation length of
256 tokens for every dataset.  The same prompts are shared by all decoding
methods. Their exact wording is given in Appendix~\ref{app:prompts}.

\paragraph{Models and Baselines.}
We evaluate LLaDA-8B-Instruct~\citep{nie2026large} and Dream-7B-Instruct~\citep{ye2025dream7bdiffusionlarge}. We compare with each model's default sampler and with Fast-dLLM~\citep{wu2026fast}, EB-Sampler~\citep{ben2026accelerated}, LoPA~\citep{xu2025lopascalingdllminference}, and DAPD~\citep{kim2026dapd}. The default samplers are LLaDA's block-wise greedy decoder and Dream's full-canvas entropy sampler (temperature 0.1, top-$p=0.9$). All accelerated baselines and RPD-block decode within 32-token blocks, whereas full RPD decodes over the full canvas. All methods share the same checkpoints, prompts, generation budget, and evaluators; method-specific settings are given in Appendix~\ref{app:experimental-details}.

\paragraph{Metrics.}
We report accuracy on GSM8K and MATH-500 and pass@1 on HumanEval and MBPP. For efficiency, we report the number of function evaluations (NFE) and throughput in tokens per second (TPS). NFE is the average number of backbone forward passes per example, where each lookahead or branch evaluation counts as a separate pass. TPS is the number of generated tokens divided by the wall-clock time of the decoding loop, measured with batch size 1 on a single NVIDIA A6000 with GPU synchronization before each timing. NFE reflects model evaluation cost, while TPS reflects practical decoding speed, including the overhead of each method's selection rule.

\begin{table*}[t]
\centering
\caption{
Main results across mathematical reasoning and code generation benchmarks.
\textbf{Bold} indicates the best result among all methods, while \underline{underline} indicates the better result between our two variants when neither is best overall.
}
\label{tab:main}

\setlength{\tabcolsep}{6pt}
\renewcommand{\arraystretch}{1.0}

\resizebox{0.9\linewidth}{!}{
\begin{tabular}{lll|ccccc|>{\columncolor{gray!12}}c>{\columncolor{gray!12}}c}
    \toprule
    \textbf{Model} 
    & \textbf{Benchmark} 
    & \textbf{Metric}
    & \textbf{Default}
    & \textbf{Fast-dLLM}
    & \textbf{EB-Sampler}
    & \textbf{LoPA}
    & \textbf{DAPD}
    & \textbf{RPD-block}
    & \textbf{RPD} \\
    \midrule
    
    \multirow{12}{*}{\shortstack[c]{Dream-7B\\Instruct}}
    & \multirow{3}{*}{HumanEval}
    & pass@1\ $\uparrow$ & 57.32 & 60.37 & 56.71 & 55.49 & 55.49 & 59.76 & \textbf{60.98} \\
    & & NFE $\downarrow$ & 256.00 & 64.54 & 78.35 & \textbf{35.79} & 77.74 & \underline{61.72} & 63.09 \\
    & & TPS $\uparrow$ & 7.35 & 31.69 & 25.57 & 19.71 & 20.93 & \textbf{31.71} & 30.07 \\
    \cmidrule(lr){2-10}

    & \multirow{3}{*}{MBPP}
    & pass@1\ $\uparrow$ & 56.80 & 58.20 & 56.80 & 51.20 & 56.40 & 58.00 & \textbf{58.40} \\
    & & NFE $\downarrow$ & 256.00 & 57.63 & 62.41 & \textbf{36.95} & 65.05 & \underline{55.49} & 55.89 \\
    & & TPS $\uparrow$ & 4.09 & 16.82 & 15.30 & 11.78 & 13.82 & \textbf{17.34} & 15.14 \\
    \cmidrule(lr){2-10}

    & \multirow{3}{*}{GSM8K}
    & Acc.\ $\uparrow$ & 81.35 & 81.73 & 81.35 & 81.65 & 80.59 & 81.27 & \textbf{81.80} \\
    & & NFE $\downarrow$ & 256.00 & 76.04 & 89.26 & \textbf{37.20} & 79.01 & \underline{70.78} & 82.03 \\
    & & TPS $\uparrow$ & 8.98 & 30.03 & 25.53 & 20.90 & 22.02 & \textbf{30.82} & 24.56 \\
    \cmidrule(lr){2-10}

    & \multirow{3}{*}{MATH-500}
    & Acc.\ $\uparrow$ & 41.80 & \textbf{46.00} & 45.00 & 44.00 & 41.80 & 44.20 & \underline{45.80} \\
    & & NFE $\downarrow$ & 256.00 & 114.34 & 124.60 & \textbf{55.28} & 112.64 & \underline{108.93} & 116.17 \\
    & & TPS $\uparrow$ & 12.57 & 29.74 & 27.09 & 20.43 & 24.07 & \textbf{30.51} & 27.71 \\

    \midrule

    \multirow{12}{*}{\shortstack[c]{LLaDA-8B\\Instruct}}
    & \multirow{3}{*}{HumanEval}
    & pass@1\ $\uparrow$ & 43.29 & 43.29 & 43.90 & 40.24 & 41.46 & 41.46 & \textbf{43.90} \\
    & & NFE $\downarrow$ & 256.00 & 44.48 & 59.06 & \textbf{24.68} & 49.09 & 41.23 & \underline{40.59} \\
    & & TPS $\uparrow$ & 7.70 & 43.13 & 32.10 & 29.32 & 36.46 & \textbf{46.23} & 45.29 \\
    \cmidrule(lr){2-10}

    & \multirow{3}{*}{MBPP}
    & pass@1\ $\uparrow$ & 39.60 & 40.20 & 39.20 & 39.40 & 37.80 & 38.60 & \textbf{40.60} \\
    & & NFE $\downarrow$ & 256.00 & 44.73 & 53.26 & \textbf{23.94} & 47.16 & 40.61 & \underline{38.16} \\
    & & TPS $\uparrow$ & 5.93 & 33.13 & 26.96 & 21.58 & 28.62 & 33.91 & \textbf{35.92} \\
    \cmidrule(lr){2-10}

    & \multirow{3}{*}{GSM8K}
    & Acc.\ $\uparrow$ & 76.50 & 76.35 & \textbf{79.15} & 76.50 & 76.19 & 75.74 & \underline{78.85} \\
    & & NFE $\downarrow$ & 256.00 & 77.09 & 88.79 & \textbf{38.35} & 75.77 & \underline{69.08} & 71.76 \\
    & & TPS $\uparrow$ & 10.50 & 35.63 & 30.72 & 25.83 & 34.34 & \textbf{38.55} & 35.30 \\
    \cmidrule(lr){2-10}

    & \multirow{3}{*}{MATH-500}
    & Acc.\ $\uparrow$ & 38.00 & 37.40 & \textbf{39.00} & 38.60 & 37.20 & 37.80 & \underline{38.80} \\
    & & NFE $\downarrow$ & 256.00 & 103.78 & 116.01 & \textbf{50.39} & 98.71 & \underline{93.99} & 98.88 \\
    & & TPS $\uparrow$ & 11.31 & 27.59 & 24.46 & 19.48 & 28.18 & \textbf{30.00} & 26.77 \\

    \bottomrule
\end{tabular}
}
\end{table*}
\subsection{Main Results}
\label{sec:main_results}

Table~\ref{tab:main} compares RPD with existing decoding methods across two models and four benchmarks. RPD variants achieve the highest throughput in every setting, with 2.4--6.1$\times$ speedups over the default samplers.

\paragraph{Accuracy.} These speedups come without a loss in accuracy. RPD ranks first or second in accuracy in all eight settings. Relative to Fast-dLLM, which commits tokens by confidence alone, RPD improves accuracy in seven settings and remains within 0.2 points in the remaining one. Only Fast-dLLM on Dream MATH-500 and EB-Sampler on the LLaDA math benchmarks exceed RPD, the latter at noticeably lower throughput. Combining layerwise stability with upstream uncertainty thus identifies reliable tokens more effectively than a single confidence or entropy criterion.

\paragraph{RPD versus RPD-block.} The two variants offer complementary trade-offs. Full RPD removes the block structure and relies on the cumulative entropy budget to order commitment. It improves accuracy over RPD-block in all eight settings at a modest cost in throughput, indicating that adaptive ordering based on upstream uncertainty is more reliable than a fixed block schedule. This supports the finding in Section~\ref{sec:upstream_uncertainty}: deferring predictions with uncertain upstream context improves reliability, while layerwise stability alone already enables aggressive parallel commitment.

\paragraph{NFE and TPS.} LoPA requires the fewest forward evaluations, but it has the lowest TPS among the accelerated methods, since its lookahead branches increase the cost of each evaluation. NFE therefore does not fully reflect decoding cost, and we treat TPS as the primary efficiency metric. RPD adds only layerwise projections to each forward pass, and its overhead remains small (Section~\ref{sec:decoding}).

\subsection{Quality--Efficiency Trade-offs}
\label{sec:tradeoff}

\begin{figure}[t]
    \centering
    \includegraphics[width=\linewidth]{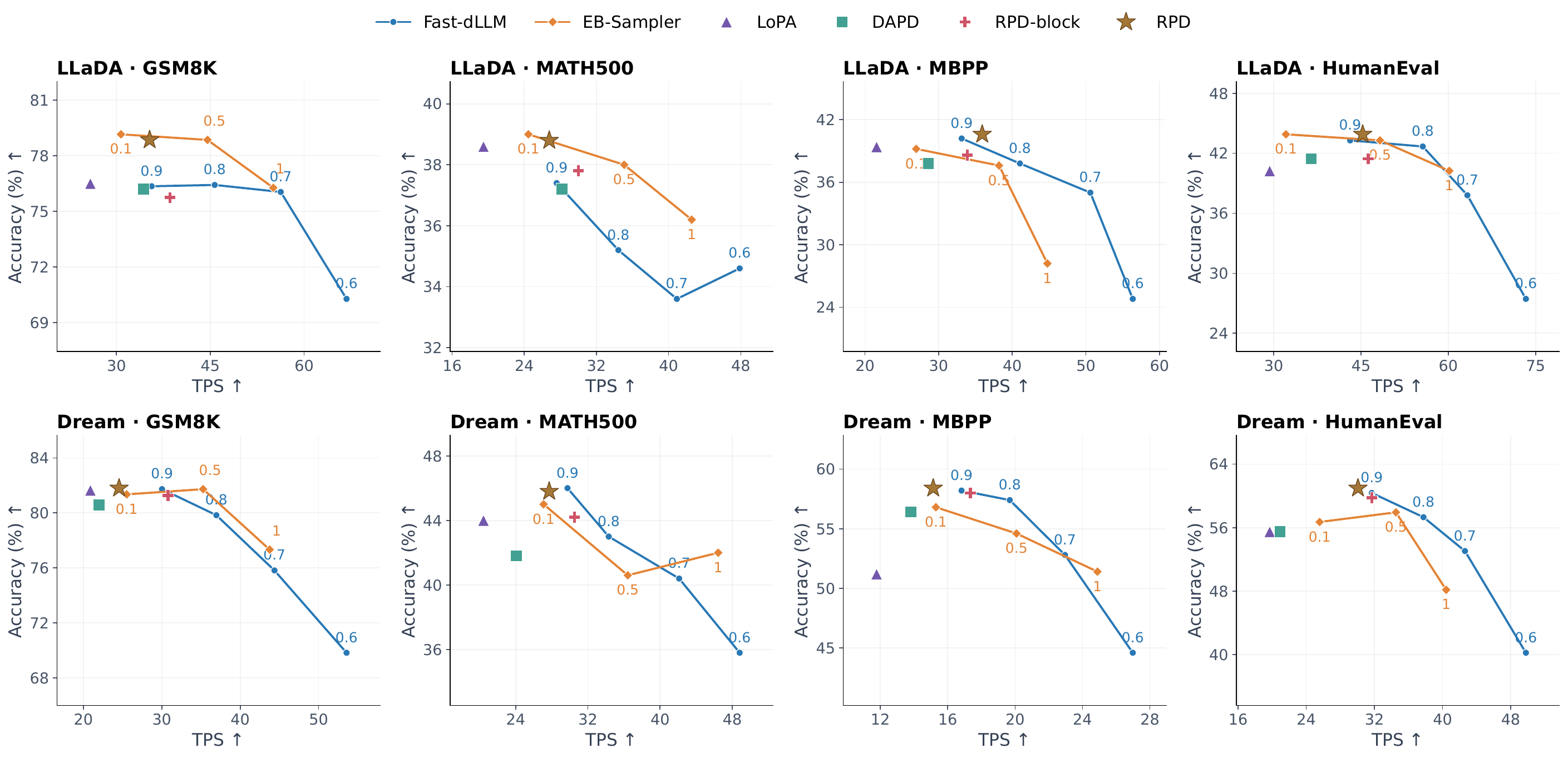}
    \caption{Accuracy--throughput trade-offs on LLaDA (top) and Dream (bottom). Fast-dLLM and EB-Sampler are swept over their confidence threshold and entropy budget, respectively, with each point labeled by its value. Other methods are shown at their default configurations.}
    \label{fig:quality-efficiency}
\end{figure}

Table~\ref{tab:main} evaluates each method at a single configuration. To examine whether the baselines can reach a better trade-off by adjusting their thresholds, Figure~\ref{fig:quality-efficiency} sweeps the confidence threshold of Fast-dLLM and the entropy budget of EB-Sampler.

\paragraph{Baseline trade-offs.} The default configurations of both baselines lie at the conservative end of their curves. Relaxing the thresholds increases throughput but reduces accuracy, and the drop is steep on code generation: on LLaDA, Fast-dLLM falls from 40.2 to 24.8 on MBPP and from 43.29 to 27.44 on HumanEval as its threshold decreases from 0.9 to 0.6. A single confidence or entropy criterion cannot distinguish reliable predictions from unsupported ones, so admitting more tokens also admits more errors (Section~\ref{sec:diagnosis}).

\paragraph{Position of RPD.} On LLaDA, RPD lies above the Fast-dLLM curve on all four benchmarks and on or above the EB-Sampler curve, with the clearest margin on MBPP. On Dream, RPD lies at the high-accuracy end of the frontier, matching the accuracy of the most conservative baseline configurations. Here the gains are smaller because confidence-based decoding already retains the accuracy of default decoding, leaving limited room for improvement. RPD-block shifts toward higher throughput at slightly lower accuracy, consistent with Section~\ref{sec:main_results}. LoPA and DAPD lie below or to the left of the frontiers in all settings.

\subsection{Ablation Studies}
\label{sec:ablation}
\begin{table*}[t]
\centering
\caption{Component ablation of RPD. Starting from confidence-based decoding, we add prediction persistence (PP), confidence drop (CD), and cumulative entropy (CE). Confidence (full canvas) commits all predictions with confidence of at least 0.9 without block structure. We report accuracy (Acc.), number of function evaluations (NFE), and throughput (TPS). On Dream, confidence decoding commits EOS tokens early and truncates the response, so few tokens count toward TPS.}
\label{tab:ablation}
\scriptsize
\setlength{\tabcolsep}{4.2pt}
\resizebox{\textwidth}{!}{
\begin{tabular}{l|ccc|ccc|ccc|ccc}
\toprule
& \multicolumn{6}{c|}{\textbf{LLaDA-8B-Instruct}}
& \multicolumn{6}{c}{\textbf{Dream-7B-Instruct}} \\
\cmidrule(lr){2-7}
\cmidrule(lr){8-13}

\textbf{Variant}
& \multicolumn{3}{c|}{\textbf{GSM8K}}
& \multicolumn{3}{c|}{\textbf{HumanEval}}
& \multicolumn{3}{c|}{\textbf{GSM8K}}
& \multicolumn{3}{c}{\textbf{HumanEval}} \\
\cmidrule(lr){2-4}
\cmidrule(lr){5-7}
\cmidrule(lr){8-10}
\cmidrule(lr){11-13}

& Acc.$\uparrow$ & NFE$\downarrow$ & TPS$\uparrow$
& Acc.$\uparrow$ & NFE$\downarrow$ & TPS$\uparrow$
& Acc.$\uparrow$ & NFE$\downarrow$ & TPS$\uparrow$
& Acc.$\uparrow$ & NFE$\downarrow$ & TPS$\uparrow$ \\
\midrule

Confidence
& 54.59 & 88.02 & 28.65
& 27.44 & 48.93 & 31.03
& 35.56 & 105.89 & 1.69
& 21.95 & 90.07 & 3.57 \\

+ block32
& 76.35 & 77.09 & 35.63
& 43.29 & 44.48 & 43.13
& 81.73 & 76.04 & 30.03
& 60.37& 64.54 & 31.69 \\

+ PP
& 74.00 & 44.80 & 57.50
& 31.10 & 28.88 & 64.57
& 71.87 & 48.41 & 43.53
& 48.78 & 47.97 & 40.06 \\

+ CE
& 78.92 & 80.84 & 33.23
& 43.29 & 45.23 & 42.46
& 81.88 & 88.03 & 24.53
& 60.98 & 65.62 & 29.75 \\

+ PP + CD
& 75.74 & 69.08 & 38.55
& 41.46 & 41.23 & 46.23
& 81.27 & 70.78 & 30.82
& 59.76 & 61.72 & 31.71 \\

\textbf{Full RPD}
& 78.85 & 71.76 & 35.30
& 43.90 & 40.59 & 45.29
& 81.80 & 82.03 & 24.56
& 60.98 & 63.09 & 30.07 \\
\bottomrule
\end{tabular}
}
\end{table*}

Table~\ref{tab:ablation} starts from confidence-based decoding over the full canvas and adds ordering constraints and the components of RPD: prediction persistence (PP, Eq.~\ref{eq:prediction_persistence}), confidence drop (CD, Eq.~\ref{eq:confidence_drop}), and cumulative entropy (CE, Eq.~\ref{eq:cumulative_entropy}). Without any ordering constraint, confidence-based decoding degrades sharply: removing the block structure lowers accuracy from 76.35 to 54.59 on LLaDA GSM8K and from 81.73 to 35.56 on Dream GSM8K. Confident predictions near the end of the sequence are committed before their supporting computations (Figure~\ref{fig:decoding}a), and on Dream these early commitments are often EOS tokens that truncate the response. Block-wise decoding avoids this failure through a fixed order. CE also decodes over the full canvas, yet matches or exceeds the accuracy of block-wise decoding in all four settings, with the largest gain on LLaDA GSM8K (+2.6 points). Deferring candidates with uncertain upstream context thus provides an adaptive commitment order without a fixed block schedule, at the cost of more forward evaluations.

Layerwise stability recovers this parallelism. Within blocks, PP alone reduces NFE by 26--42\% but causes large accuracy drops, for example from 43.29 to 31.10 on LLaDA HumanEval, because it admits predictions that keep the same token across layers while losing probability. Adding CD removes most of these predictions, recovering most of the lost accuracy while requiring fewer forward evaluations than block-wise confidence decoding in all four settings (Table~\ref{tab:ablation}). Combining the two, full RPD retains the accuracy of CE while reducing NFE by 4--11\% and matching or increasing throughput in all four settings.





\subsection{Decoding Behavior and Overhead}
\label{sec:decoding}

\begin{figure}[t]
    \centering
    \includegraphics[width=\linewidth]{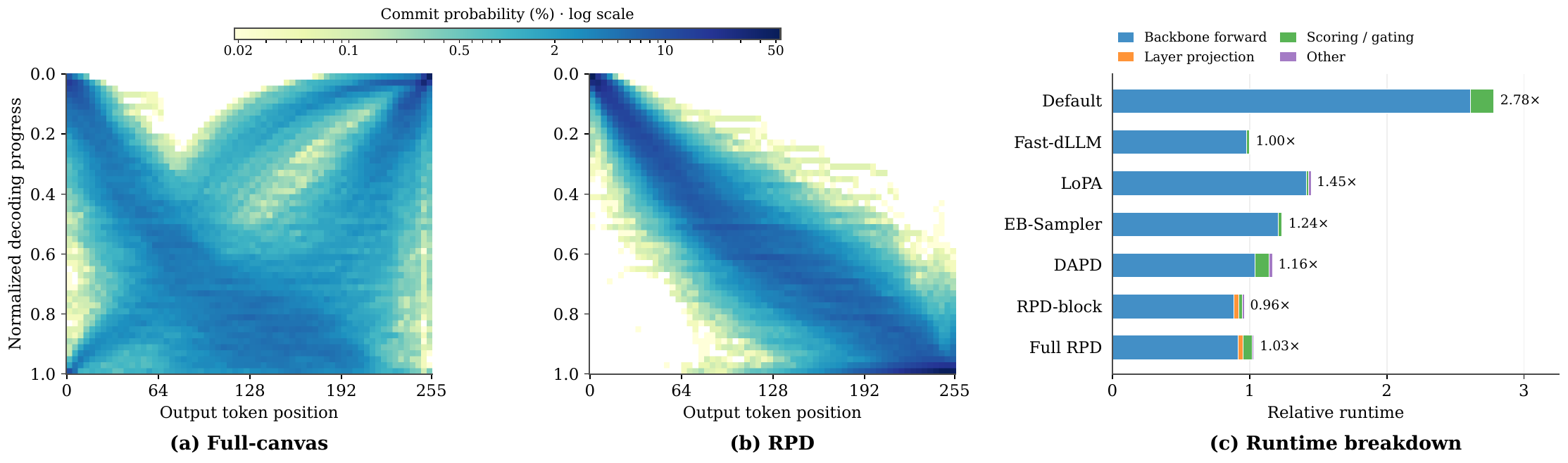}
    \caption{(a, b) Commitment order of Full-canvas and RPD on LLaDA GSM8K. Each cell shows the percentage of samples in which an output position is committed at a given normalized decoding progress. (c) Runtime of each method relative to Fast-dLLM, decomposed into backbone forward, layer projection, scoring and gating, and other operations.}
    \label{fig:decoding}
    \vspace{-1pt}
\end{figure}

\paragraph{Commitment order.}
Figure~\ref{fig:decoding}a--b shows when each output position is committed on LLaDA GSM8K. Full-canvas commits both ends of the sequence early and fills the middle last, producing an X-shaped pattern. Positions near the end, where the final answer is usually written, are often committed before the reasoning that supports them, which is the premature commitment illustrated in Figure~\ref{fig:intro}. RPD also decodes over the full canvas, yet commits positions in an approximately left-to-right order. The cumulative entropy budget thus recovers a dependency-respecting order without a fixed block schedule. Per-sample trajectories are provided in Appendix~\ref{app:trajectories}.

\paragraph{Runtime overhead.}
Figure~\ref{fig:decoding}c decomposes the runtime of each method. The backbone forward pass dominates in all cases. Layerwise projection and scoring add only a small cost, so RPD runs at 1.03$\times$ the runtime of Fast-dLLM and RPD-block at 0.96$\times$. In comparison, LoPA, EB-Sampler, and DAPD require 1.45$\times$, 1.24$\times$, and 1.16$\times$, respectively.
\section{Related Work}
\label{sec:related_work}

\paragraph{Parallel decoding in masked diffusion language models.}
MDLM~\citep{sahoo2024simple} develops masked diffusion for language modeling, while LLaDA~\citep{nie2026large} and Dream~\citep{ye2025dream7bdiffusionlarge} demonstrate its use in instruction following, mathematical reasoning, and code generation.
Fast-dLLM~\citep{wu2026fast} combines approximate KV caching with confidence thresholds for parallel decoding.
EB-Sampler~\citep{ben2026accelerated} controls parallel unmasking through an entropy budget supported by an analysis of sampling error.
Other methods search over decoding decisions.
LoPA~\citep{xu2025lopascalingdllminference} uses parallel lookahead to find token-filling orders that allow more parallel decoding in later steps.
Ripple-Pivot Search~\citep{ye2026ripplepivotsearchactiveparallel} searches for pivot positions and token assignments that reduce uncertainty at other masked positions.
Unlike these search-based methods, we assess which tokens to commit using upstream uncertainty and layerwise predictions from the current forward pass, without exploring alternative continuations.

\paragraph{Dependencies and commitment order.}
DAPD~\citep{kim2026dapd} selects independent sets from an attention-based dependency graph to avoid jointly updating strongly coupled positions.
DOS~\citep{zhou2026dependency} orders masked positions by their dependence on observed context.
\citet{yeom2026answerfirstreasonlater} show that committing answers too early can reduce reasoning accuracy and studies frontier gating to address this problem.
We also study premature commitment, but distinguish dependencies that require further decoding steps from those that can be resolved within one forward pass.
This distinction motivates separate checks for upstream uncertainty and prediction stability, allowing parallel commitment when further decoding steps are unnecessary.

\paragraph{Layerwise predictions and stability.}
The tuned lens~\citep{belrose2023eliciting} uses learned affine probes to obtain vocabulary distributions from intermediate layers, while DoLa~\citep{chuang2024dola} contrasts predictions across layers to improve factuality.
Stability-Weighted Decoding~\citep{wu2026stability} uses distributional changes between consecutive denoising steps to score tokens.
We instead examine prediction consistency across layers within one forward pass and combine it with final confidence to select tokens for commitment.
We use layerwise predictions to choose positions, without training probes or modifying token distributions through layer contrast.
\section{Conclusion}

We studied when masked diffusion language models can reliably commit multiple tokens in parallel. Our diagnostics show that confidence alone does not determine a reliable commitment order: downstream predictions are less reliable when their upstream information remains uncertain, and predictions that remain stable across layers are more likely to be correct. Based on these findings, we proposed Reliable Parallel Decoding (RPD), a training-free method that selects candidates by layerwise stability and orders their commitment by a cumulative entropy budget. Without a fixed block schedule, RPD commits tokens in a dependency-respecting order and achieves the highest throughput among the evaluated methods while maintaining or improving accuracy. 

\bibliography{iclr2027_conference}
\bibliographystyle{iclr2027_conference}

\clearpage
\appendix
\section{Appendix}

\subsection{Prompts}
\label{app:prompts}
\begin{figure}[h]\centering
\includegraphics[width=\linewidth]{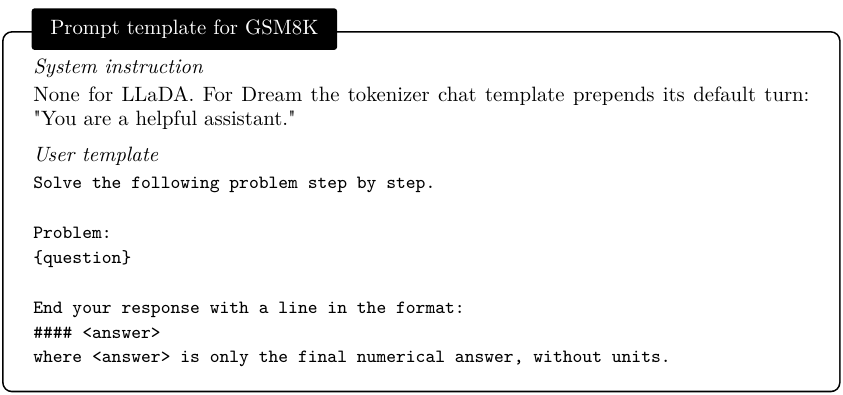}
\caption{GSM8K instruction.}\label{fig:prompt_gsm8k}\end{figure}

\begin{figure}[h]\centering
\includegraphics[width=\linewidth]{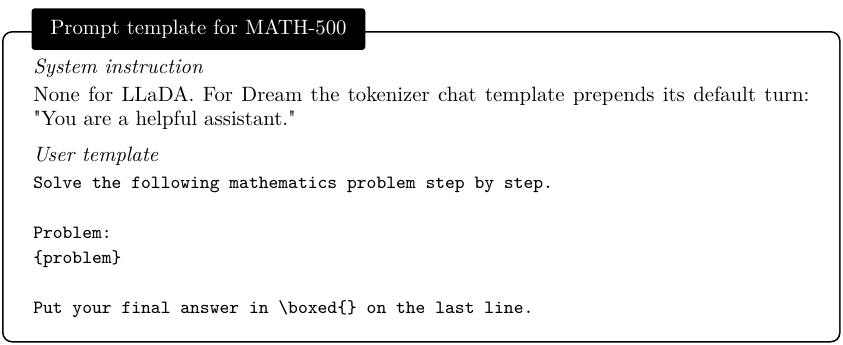}
\caption{MATH-500 instruction.}\label{fig:prompt_math500}\end{figure}

\begin{figure}[h]\centering
\includegraphics[width=\linewidth]{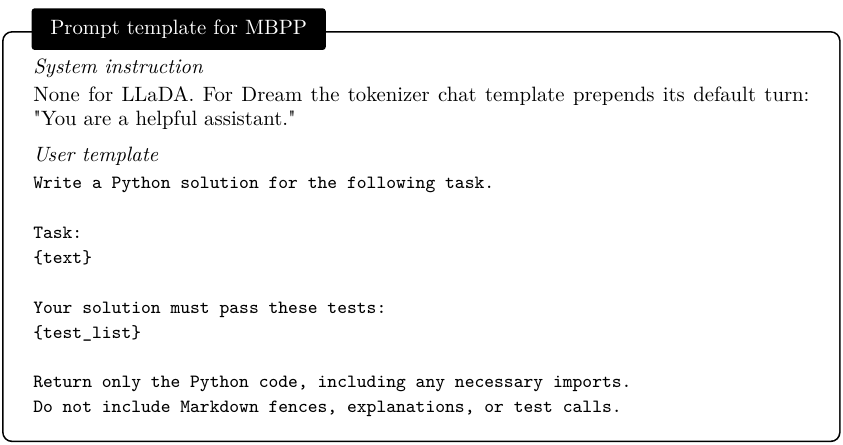}
\caption{MBPP instruction. \texttt{\{test\_list\}} is filled with the first three public tests.}\label{fig:prompt_mbpp}\end{figure}

\begin{figure}[h]\centering
\includegraphics[width=\linewidth]{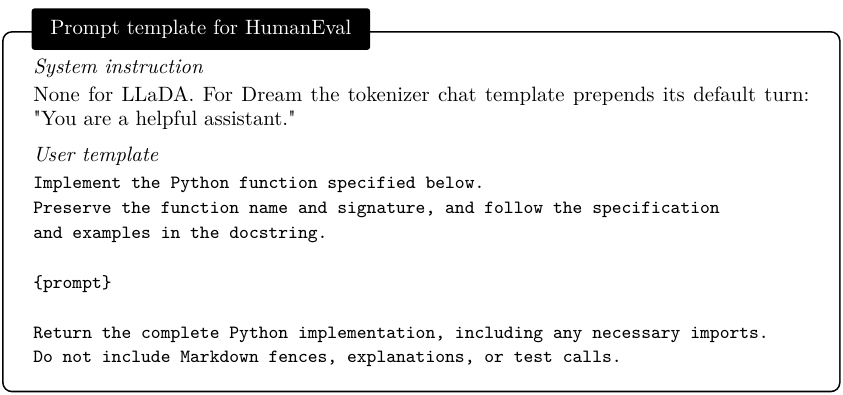}
\caption{HumanEval instruction.}\label{fig:prompt_humaneval}\end{figure}

\begin{figure}[h]\centering
\includegraphics[width=\linewidth]{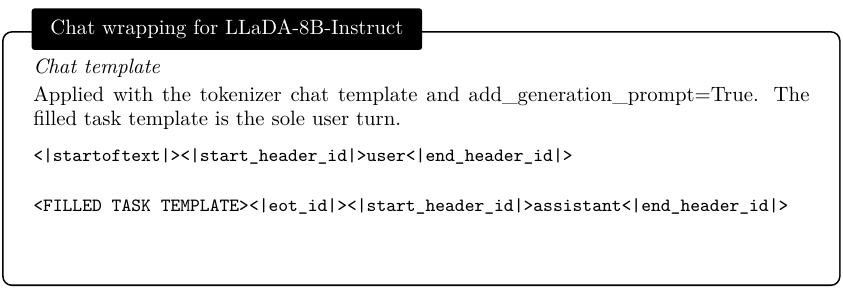}
\caption{Chat wrapping for LLaDA-8B-Instruct.}\label{fig:prompt_chat_llada}\end{figure}

\begin{figure}[h]\centering
\includegraphics[width=\linewidth]{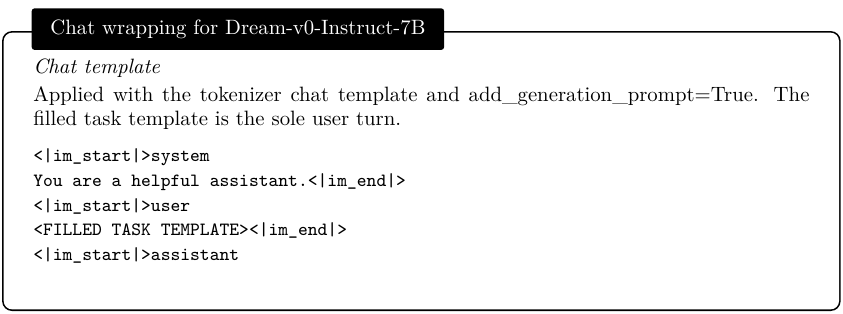}
\caption{Chat wrapping for Dream-v0-Instruct-7B.}\label{fig:prompt_chat_dream}\end{figure}

All benchmarks use a single zero-shot instruction, filled with the raw benchmark
field and wrapped in the model's native chat template with
\texttt{add\_generation\_prompt=True}; the filled instruction is the only user
turn.  Figures~\ref{fig:prompt_gsm8k}--\ref{fig:prompt_humaneval} give the four
instructions verbatim, and Figures~\ref{fig:prompt_chat_llada}--\ref{fig:prompt_chat_dream}
give the chat wrapping of each backbone.  LLaDA receives no system turn; Dream's
tokenizer template prepends its own default turn, \texttt{You are a helpful
assistant.}  We add no system message, demonstration, assistant prefill, or
input truncation beyond this.

\subsection{Diagnostic Data and Implementation}
\label{app:diagnostic_setup}

\paragraph{Data construction.}
We deterministically sample 2,000 examples from the GSM8K-Aug training set using seed 325.
Each example contains a verified multi-step arithmetic trace and an integer final answer.
We require at least 48 eligible upstream digit positions and exclude examples in which the final answer appears literally in the question, an earlier equation, or the left-hand side of the final equation.

The input consists of the question followed by the gold reasoning trace, with the right-hand side of the final equation serving as the target answer.
For each example, we independently sample 16 upstream masking patterns for each mask count $m\in\{4,8,16,32\}$.
Selected upstream digits and all target-answer digits are masked simultaneously; all other positions retain their gold tokens.
This produces 64 states per question and 128,000 states per model.
LLaDA and Dream use identical questions and character-level masking patterns, mapped into each tokenizer through explicit digit tokenization.
All experiments in this paper use the public instruction-tuned checkpoints
\texttt{GSAI-ML/LLaDA-8B-Instruct} (revision \texttt{08b83a6}) and
\texttt{Dream-org/Dream-v0-Instruct-7B} (revision \texttt{05334cb}), in bfloat16.
Masking patterns are defined over characters so that both backbones see the
same masks.  To map them into each tokenizer we first verify that every digit
\texttt{0}--\texttt{9} encodes to exactly one token in that vocabulary, then
build the input by explicit segmentation: the reasoning prefix is encoded in
runs, and each digit character is emitted as its own single-token encoding,
recording the resulting token index for that character.  Digits are masked only
inside arithmetic content, never in step-number labels, and every target-answer
digit is emitted the same way.  A masking pattern therefore selects the same
characters, and hence positionally corresponding tokens, in both models; we
assert that all selected upstream positions precede the first target position.

\paragraph{Prediction and scoring.}
Every masked state is evaluated with one independent forward pass.
Diagnostic probabilities are taken before temperature scaling, sampling noise, or top-$p$ truncation.
For the upstream-entropy analysis, a state is correct only if the complete decoded numerical answer matches the gold answer.
For the layerwise analysis, a target token is correct if its final-layer argmax matches the gold token.
There are 531,264 target-token observations per model.

\paragraph{Intermediate-layer readout.}
We apply the model's final normalization and language-model head to intermediate hidden states.
At the final layer, we use native logits without applying final normalization twice.
Dream predictions follow its native one-position logit alignment.
The trajectory analysis uses the final half of the transformer layers.
Concretely, this is layers $17$--$32$ of LLaDA's $32$ layers and layers
$15$--$28$ of Dream's $28$ layers, in both cases the final layer together with
the posterior half of the stack, giving $16$ and $14$ trajectory points
respectively.

Let $y_i$ denote the final-layer argmax at position $i$.
Scanning backward from the final layer, we count consecutive layers predicting $y_i$ until the first disagreement or the beginning of the analysis window.
This count is $K_i$, and the suffix starts at $s_i=L-K_i+1$.
Earlier matches separated by a disagreement are excluded.
The peak-to-final drop $r_i$ is computed from the probability of the same token $y_i$ throughout this suffix.

\subsection{Complete Diagnostic Results}
\label{app:diagnostic_results}

\subsubsection{Upstream Entropy}
\label{app:upstream_results}

For each model and mask count, we retain questions whose 16 masking patterns contain both correct and incorrect answers.
Within each retained question, patterns are ranked by cumulative upstream entropy.
We average answer-error labels over the five lowest-entropy patterns and separately over the five highest-entropy patterns.
The reported rates average these question-level values with equal weight.
We compute 95\% confidence intervals using 2,000 question-level bootstrap resamples, preserving the low--high pairing within each question.
The comparison does not match configurations by target confidence or mask distance.

Table~\ref{tab:upstream_complete} reports all groups.
High-entropy configurations have higher error rates for both models at every mask count, and all paired differences have 95\% confidence intervals above zero.

\begin{table}[t]
\centering
\small
\caption{Complete upstream-entropy results. $n$ is the number of retained mixed-outcome questions. Error rates are percentages; $\Delta$ is the paired high-minus-low difference in percentage points, computed before rounding.}
\label{tab:upstream_complete}
\begin{tabular}{llrrrr}
\toprule
Model & Masks & $n$ & Low entropy & High entropy & $\Delta$ \\
\midrule
LLaDA & 4  & 40   & 22.5 & 51.0 & 28.5 \\
      & 8  & 81   & 14.6 & 38.8 & 24.2 \\
      & 16 & 182  & 12.7 & 37.6 & 24.8 \\
      & 32 & 813  & 10.5 & 39.0 & 28.6 \\
\midrule
Dream & 4  & 128  & 15.8 & 33.1 & 17.3 \\
      & 8  & 322  & 9.3  & 27.2 & 17.9 \\
      & 16 & 762  & 9.4  & 27.7 & 18.3 \\
      & 32 & 1513 & 21.1 & 51.0 & 29.9 \\
\bottomrule
\end{tabular}
\end{table}

Paired differences with 95\% question-level bootstrap intervals, in percentage
points, are $28.5$ $[18.0, 38.5]$, $24.2$ $[18.0, 29.9]$, $24.8$ $[21.0, 28.9]$
and $28.6$ $[26.5, 30.5]$ for LLaDA, and $17.3$ $[13.9, 20.9]$, $17.9$
$[15.5, 20.2]$, $18.3$ $[16.4, 20.2]$ and $29.9$ $[28.4, 31.5]$ for Dream, at
$4$, $8$, $16$ and $32$ masks respectively.  Every interval lies above zero.

\subsubsection{Layerwise Agreement and Probability Drop}
\label{app:layerwise_results}

The agreement curves group target tokens by final confidence into
$[0.6,0.7)$, $[0.7,0.8)$, $[0.8,0.9)$, $[0.9,0.95)$, and $[0.95,1]$.
For each interval and threshold $k\in\{2,\ldots,10\}$, we report the token error rate among observations satisfying $K_i\ge k$.
These selections are nested, and empty selections are omitted.
The curves are token weighted.

The Dream heatmap includes observations with final confidence in $[0.6,0.9)$, $K_i\ge2$, and $r_i\ge0.08$.
Rows group agreement lengths into $2$, $3$--$4$, $5$--$8$, and $\ge9$.
Columns use drop intervals $[0.08,0.16)$, $[0.16,0.24)$, $[0.24,0.32)$, and $[0.32,0.40]$.
Each cell reports the token-weighted error rate and observation count.

Two patterns follow.  First, within a fixed confidence interval, requiring a
longer agreement suffix lowers the error rate, and the effect is largest exactly
where the decision is hardest: in the $[0.6,0.7)$ interval Dream's error rate
falls from $26.5\%$ at $K_i\ge2$ to $4.6\%$ at $K_i\ge10$, while in
$[0.95,1]$ the curve is already flat, because such predictions are reliable
whether or not they persist across layers.  This is what motivates applying the
stability test only below $\theta_h$.  Second, the two quantities interact: a
large confidence drop is tolerable when the prediction has persisted for many
layers but not when it has just appeared, which is why the score combines them
as $\min(K_i,K_{\max}) - w\,r_i$ rather than thresholding either alone.

\subsection{Commitment Order under Model-Specific Decoding}
\label{app:commitment_order}

All traces in this section commit exactly one position per forward pass on a
256-position canvas, for the first GSM8K document, so the panels differ only in
the setting named in their title.

\begin{figure}[h]\centering
\includegraphics[width=\linewidth]{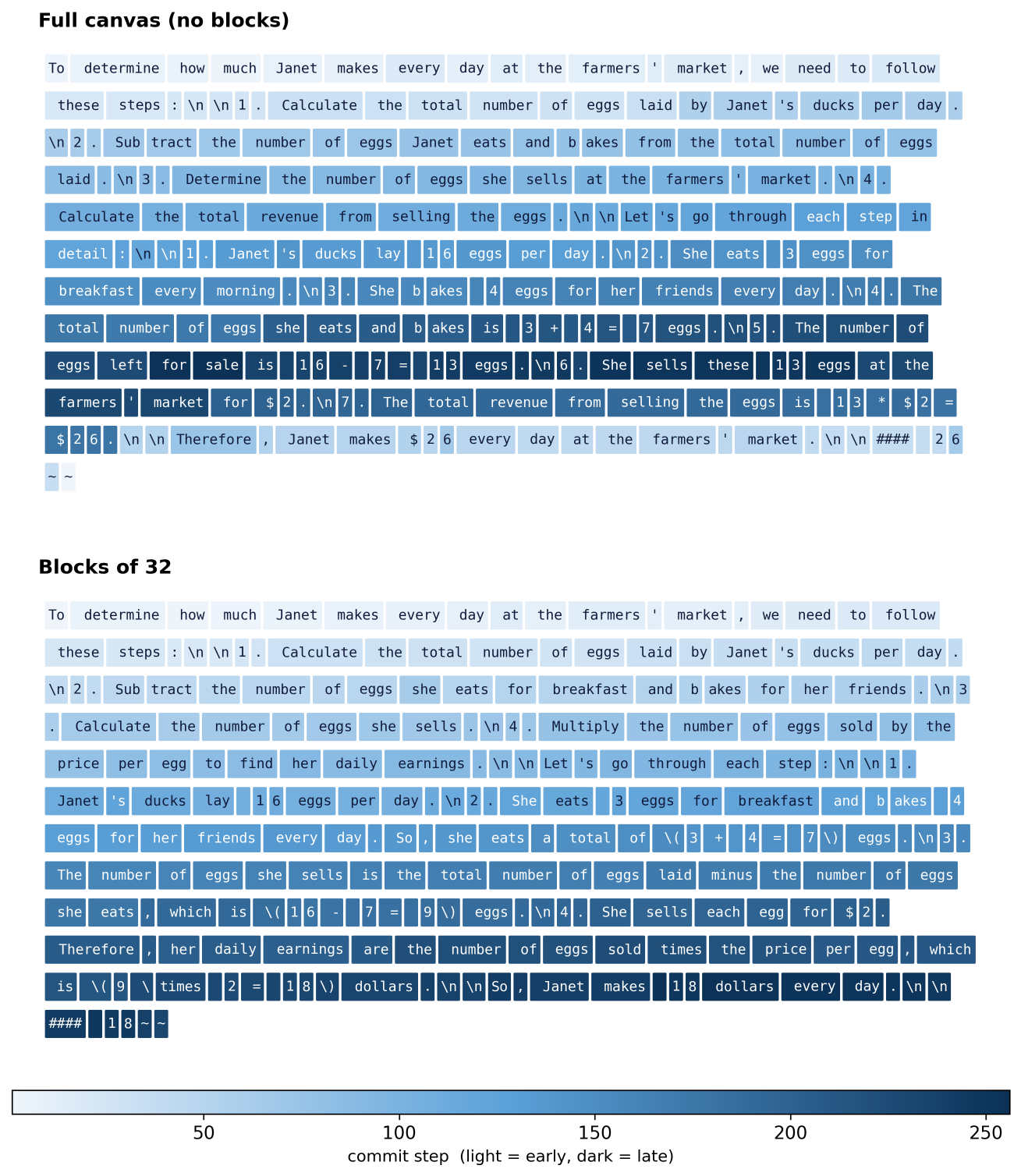}
\caption{LLaDA commitment order with and without block structure, one commit per forward pass. Each box is one generated token, shaded by the step at which it was committed.}\label{fig:traj_llada}\end{figure}

\paragraph{LLaDA: with and without blocks.}
LLaDA's block-wise decoder processes blocks sequentially, preventing commitment
in later blocks before the active block is completed.  Figure~\ref{fig:traj_llada}
contrasts this with the same greedy maximum-confidence rule applied to the full
canvas.  With blocks of 32 the commitment order is a clean left-to-right
diagonal.  Without blocks the order is still broadly left-to-right, but the
second half of the canvas receives scattered late commits that interleave with
positions committed much earlier: whole spans of the arithmetic are still dark
while the text around them is already light.  The block schedule, not the confidence rule,
is what enforces the order.  This trace reproduces the archived default run
token for token.

\begin{figure}[h]\centering
\includegraphics[width=\linewidth]{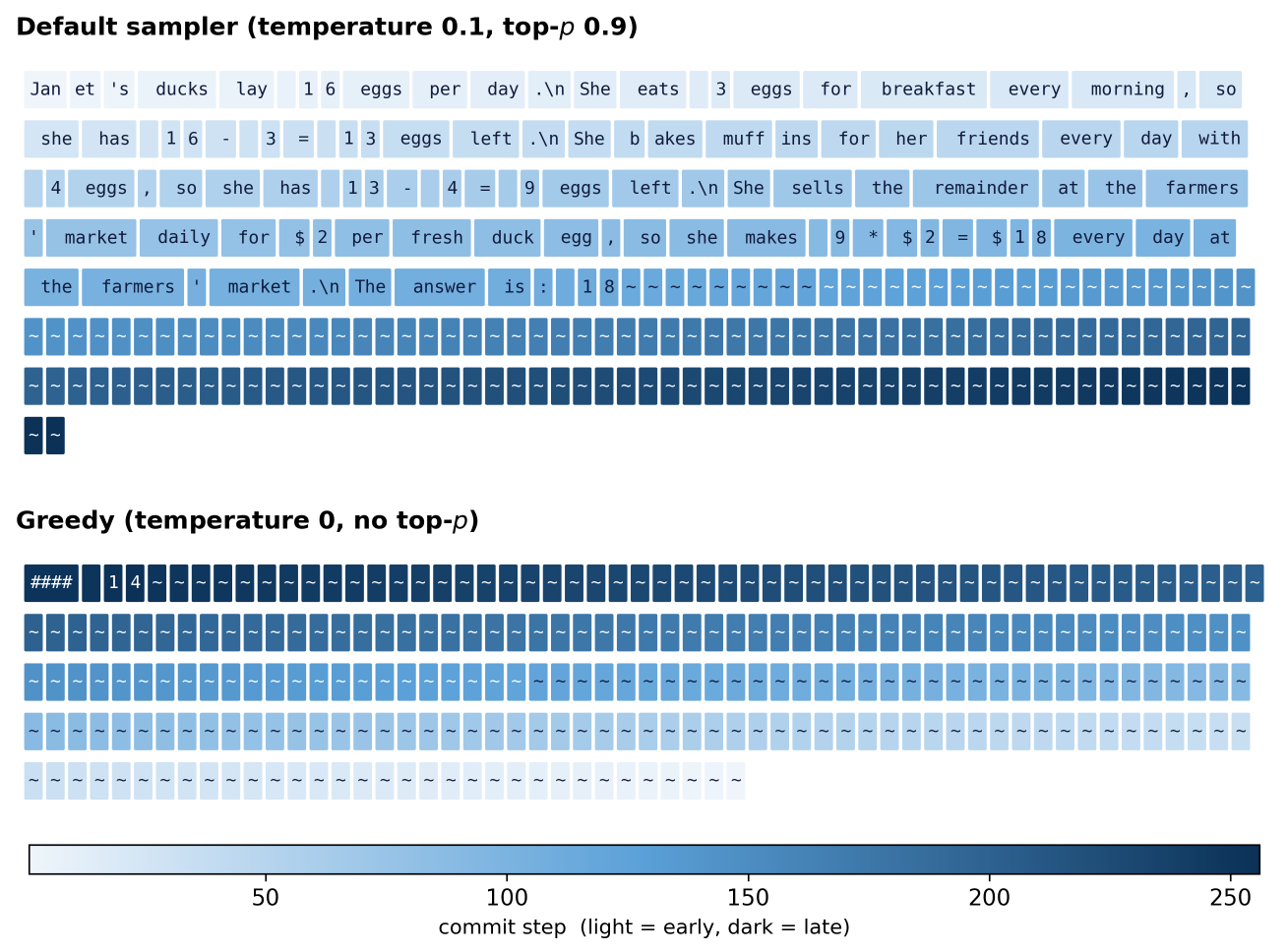}
\caption{Dream commitment order under its default sampler and under greedy decoding, one commit per forward pass and lowest-entropy-first selection in both panels. Each box is one generated token, shaded by the step at which it was committed; \texttt{\textasciitilde} marks a terminal token.}\label{fig:traj_dream}\end{figure}

\paragraph{Dream: the role of the default sampler.}
Dream has no block structure; its default sampler draws from the distribution
obtained after temperature $0.1$ and top-$p$ $0.9$, and commits the position
whose transformed distribution has the lowest entropy.  Under this
configuration we observe an approximately left-to-right commitment order
(Figure~\ref{fig:traj_dream}, top).

The temperature is not an incidental sampling detail.  Setting it to zero, with
no top-$p$ truncation, and keeping the same lowest-entropy rule reverses the
order (Figure~\ref{fig:traj_dream}, bottom): the trailing terminal positions have
the lowest entropy on the raw distribution, so they are committed first and the
canvas fills from right to left: in Figure~\ref{fig:traj_dream} the terminal
tokens are the lightest of the panel and the only content left is the three
tokens of a truncated answer.  The
ordering score must therefore be read from the same transformed distribution the
token is drawn from.

\subsection{Detailed Experimental Configuration}
\label{app:experimental-details}

\paragraph{Prompts and evaluation.}
All experiments use dataset-specific zero-shot instructions and the model's
native chat template, without demonstrations, a system message, assistant
prefill, or input truncation.  The GSM8K prompt asks for step-by-step reasoning
and requires a final line of the form \texttt{\#\#\#\# <answer>}.  The
MATH-500 prompt similarly requests a derivation followed by a
\texttt{\textbackslash boxed\{\}} answer.  For MBPP, the prompt supplies the
task and its public tests and requests Python code only.  For HumanEval, it
supplies the canonical function signature and docstring and requests a complete
implementation without Markdown fences, explanations, or test calls.

GSM8K responses are scored by extracting the final answer and performing exact
numeric comparison, treating equivalent decimal representations identically.
MATH-500 uses symbolic answer equivalence.  HumanEval and MBPP are evaluated by
executing the generated Python implementation against their tests; we apply the
same formatting-only code-fence removal to every method.  NFE counts actual
model forward calls, including LoPA branch evaluations.  TPS excludes model
loading, tokenization, and task scoring.  Generation is synchronized before
timing, and each throughput run occupies one NVIDIA A6000 GPU.

\paragraph{Default decoding.}
Every default run commits one token per forward evaluation.  LLaDA uses greedy
maximum-confidence decoding with blocks of 32.  Dream uses its official
full-canvas entropy sampler with temperature $0.1$ and top-$p=0.9$.  Both
models generate on a 256-position canvas.

\subsubsection{Matched baseline configuration.}
\label{sec:baseline_config}
For a controlled comparison, accelerated baselines use blocks of 32 and the
same 256-token budget.

\textit{Fast-dLLM} commits, in every forward pass, all masked positions in the
active block whose final confidence exceeds $0.9$, and otherwise the single most
confident position.  The threshold is applied per position and independently of
how many positions are already committed in the same pass.  Our evaluation
disables KV caching so that NFE and token-selection effects can be compared
directly.

\textit{EB-Sampler} sorts the masked positions of the active block by an error
proxy (confidence, in our runs) and commits the longest prefix of that order
whose marginal entropies satisfy $\sum_{i\in U} H_i - \max_{i\in U} H_i \le
\gamma$, with $\gamma = 0.1$ nats.  The budget is shared by the whole prefix, so
the least certain admitted position is exempt and every additional position
consumes budget.  A single position is therefore always affordable.

\textit{LoPA} keeps a confidence anchor that commits all positions above $0.9$,
and additionally spawns three lookahead branches, each extending the anchor by
one greedy token at a distinct high-scoring position.  All four branches are
verified in one batched forward pass and scored by the mean negative entropy of
the remaining masked positions of the block; the winning branch is retained and
its logits are reused in the next iteration.  Every branch microbatch
contributes to NFE.

\textit{DAPD} uses the direct variant, which derives a dependency graph over the
masked positions from self-attention and commits an independent set of that
graph, with the official task-specific dependency thresholds.  Unless explicitly
specified above for the native Dream sampler, matched accelerated baselines use
greedy sampling without top-$p$ truncation.

\subsection{Implementation Details of RPD}
\label{app:rpd_impl}

\paragraph{Hyperparameters.}
Table~\ref{tab:rpd_hparams} lists the hyperparameters of RPD.  All values are
shared across benchmarks.  The stability threshold $\theta_s$ and the
confidence-drop weight $w$ are the only model-specific values; they were set
once per backbone and then frozen for every benchmark.

\paragraph{Choice of confidence thresholds.}
The thresholds $\theta_c$ and $\theta_h$ follow the diagnostics in Figure~\ref{fig:diagnosis}b. Predictions with final confidence of at least 0.9 have near-zero error rates regardless of prediction persistence, so a stability test adds little information for them. We therefore set $\theta_h = 0.9$ and admit such predictions directly. For lower confidence, persistence becomes informative. In the lowest analyzed
range, $[0.6, 0.7)$, Dream's token error rate decreases from $26.5\%$ at
$K_i\ge2$ to $4.6\%$ at $K_i\ge10$, reaching levels comparable to those of
higher-confidence predictions. We therefore set $\theta_c = 0.6$ and allow predictions in $[0.6, 0.9)$ to become candidates when they pass the stability test. Predictions below $\theta_c$ are never admitted in the current iteration.

\paragraph{Choice of the stability threshold and drop weight.}
Unlike $\theta_c$ and $\theta_h$, which are read directly off the confidence
axis of the diagnostics, $\theta_s$ and $w$ describe how the two layerwise
quantities should be traded against each other, and we set them from the same
analysis in Section~\ref{sec:layerwise_stability}.  Two observations fix their form.
First, the benefit of persistence saturates: beyond a handful of agreeing layers
the error rate is already flat, so the score caps persistence at $K_{\max}=6$
and $\theta_s$ only has to separate predictions that have settled from those
that have not.  Second, the tolerable confidence drop depends on how long the
prediction has persisted; for $K_i=5$--$8$ the error rate rises from $4.6\%$ to
$23.7\%$ across the drop range, while for $K_i\ge9$ the same range costs only a
few points.  A single linear penalty $w\,r_i$ reproduces this coupling: under a
fixed threshold, a longer suffix buys tolerance for a larger drop, up to the
cap.  We chose $w$ so that the implied ceiling $(K_{\max}-\theta_s)/w$ sits just
above the drop region that Figure~\ref{fig:diagnosis}c still shows as low risk,
giving $0.167$ for LLaDA and $0.175$ for Dream.  The two backbones need slightly
different values because their trajectories saturate at different depths, which
is also why $\ell_0$ differs.  The resulting pair was then frozen for both
models, all four benchmarks and both variants; no benchmark-specific or
task-specific tuning is performed anywhere in the paper.
Figure~\ref{fig:sp_drop} confirms that the committed predictions sit well inside
the implied ceiling.

\begin{table}[h]
\centering
\caption{Hyperparameters of RPD.}
\label{tab:rpd_hparams}
\begin{tabular}{llc}
\toprule
\textbf{Symbol} & \textbf{Description} & \textbf{Value} \\
\midrule
$\theta_h$ & High-confidence threshold (bypasses the stability test) & 0.9 \\
$\theta_c$ & Minimum confidence for stability-based admission & 0.6 \\
$\theta_s$ & Stability threshold & 3.5 (LLaDA) / 2.5 (Dream) \\
$K_{\max}$ & Cap on prediction persistence & 6 \\
$w$ & Weight of the confidence-drop penalty & 15 (LLaDA) / 20 (Dream) \\
$\beta$ & Cumulative entropy budget & 4 nats \\
$\ell_0$ & First layer used for layerwise projection & 17 (LLaDA) / 15 (Dream) \\
$W$ & Fallback window when no candidate is committable & 32 \\
\bottomrule
\end{tabular}
\end{table}

\paragraph{Layerwise projection.}
We compute $K_i$ and $r_i$ by projecting the hidden states of layers $\ell_0, \dots, L$ through the model's final normalization and output head, following the procedure in Section~\ref{sec:layerwise_stability}. All projections reuse the hidden states of the current forward pass and require no additional forward evaluation. Two restrictions keep this inexpensive.  First, only the currently masked
positions are projected, never the already committed prefix.  Second, only the
posterior half of the stack is projected, and the final layer reuses the raw
logits of the forward pass rather than being normalized a second time.  The
projection is further evaluated lazily, so that layers are materialized only
until the agreement suffix of a position is determined. Figure~5c reports the resulting runtime overhead.

\begin{figure}[h]\centering
\includegraphics[width=\linewidth]{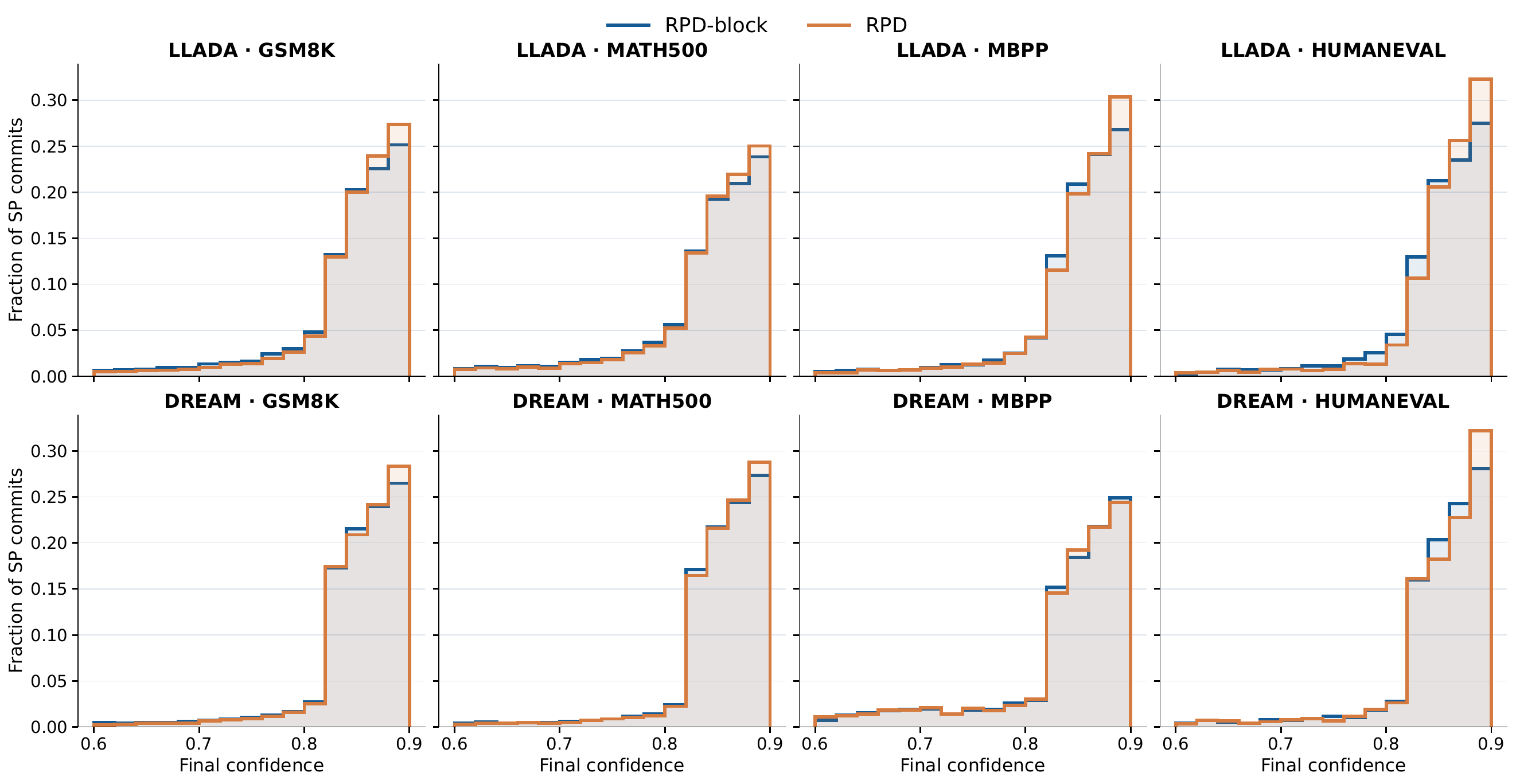}
\caption{Final confidence of tokens admitted by the stability test.}\label{fig:sp_conf}\end{figure}

\begin{figure}[h]\centering
\includegraphics[width=\linewidth]{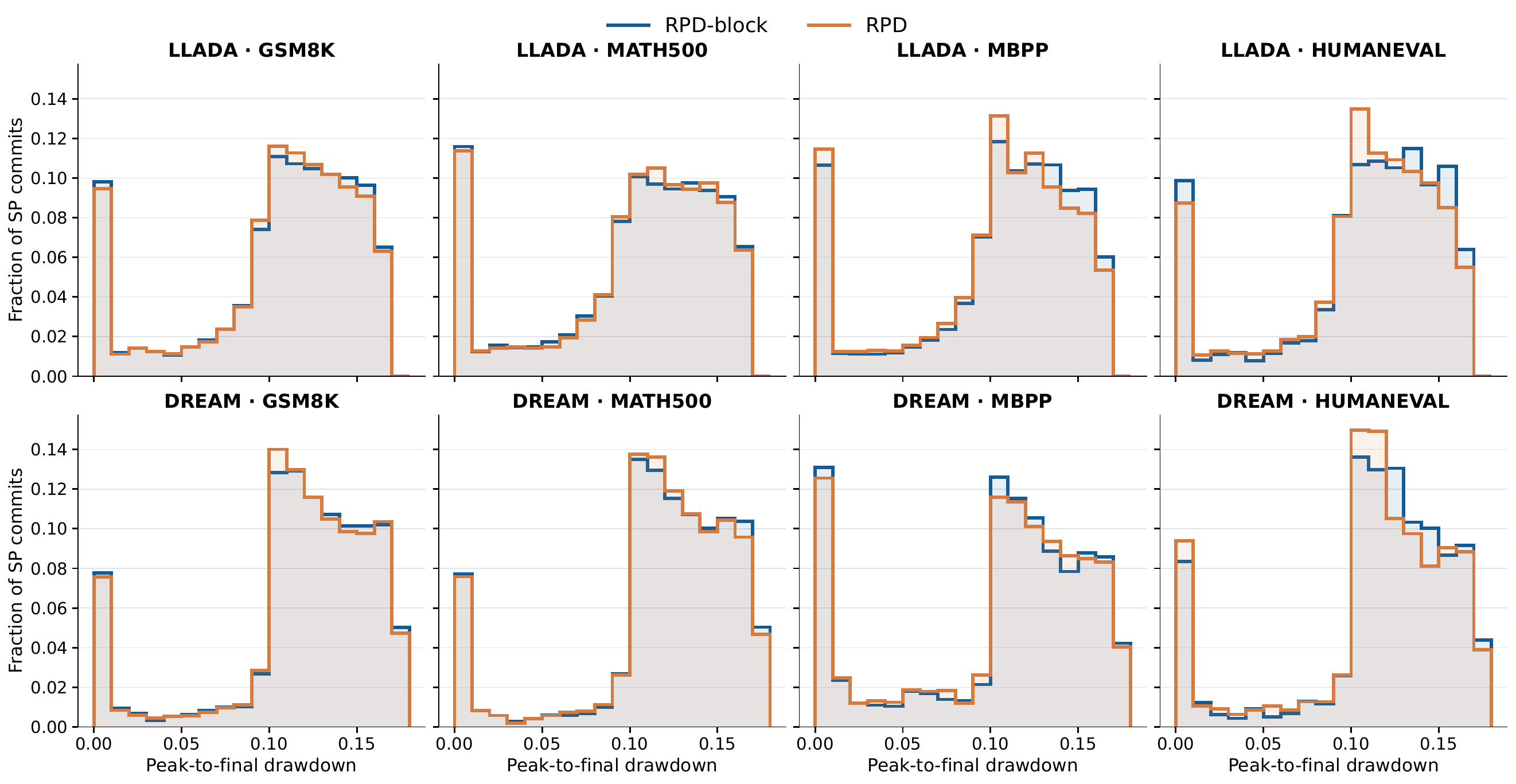}
\caption{Confidence drop $r_i$ of tokens admitted by the stability test. The upper limit in each panel is the selection ceiling $(K_{\max}-\theta_s)/w$.}\label{fig:sp_drop}\end{figure}

\paragraph{What the stability test actually admits.}
Because the admission rule is $S_i = \min(K_i, K_{\max}) - w\,r_i \ge \theta_s$
with $K_{\max}=6$, the thresholds impose a hard ceiling on the tolerated
confidence drop: $r_i \le (6-\theta_s)/w$, which is $1/6 \approx 0.167$ for
LLaDA and $0.175$ for Dream.  The distributions below are therefore
selection-induced and are not unconditional model statistics.
Figures~\ref{fig:sp_conf}--\ref{fig:sp_drop} show them over all committed
stability-admitted tokens of the full runs.  On LLaDA GSM8K, RPD-block admits
$29{,}957$ tokens through the stability route against $273{,}423$ through the
high-confidence route and $34{,}284$ by fallback; the admitted tokens have
median confidence $0.858$ $[0.774, 0.892]$ at the $10$th and $90$th percentiles,
median persistence $K_i=7$, and median drop $0.117$ with a $90$th percentile of
$0.156$.  Full RPD on the same cell is nearly identical
($27{,}991$ stability commits, median confidence $0.861$, median drop $0.116$),
and Dream behaves the same way at its own ceiling (median drop $0.125$,
$90$th percentile $0.165$).  No committed token exceeds a drop of $0.25$ in any
cell.  The stability route therefore contributes roughly $8$--$10\%$ of all
commits, concentrated in the intermediate confidence band it was designed for.

\paragraph{RPD-block.}
RPD-block applies the candidate selection of
Section~\ref{sec:method_stability} within fixed decoding blocks of 32 tokens and
omits the cumulative entropy budget, relying on the block order alone.  It uses
the same $\theta_h$, $\theta_c$, $\theta_s$, $K_{\max}$, $w$ and $\ell_0$ as
Full RPD; only the commitment rule differs.  When no candidate passes, it falls
back to the most confident masked position of the active block.

\subsection{Trajectories Visualization}
\label{app:trajectories}

\begin{figure}[h]\centering
\includegraphics[width=\linewidth]{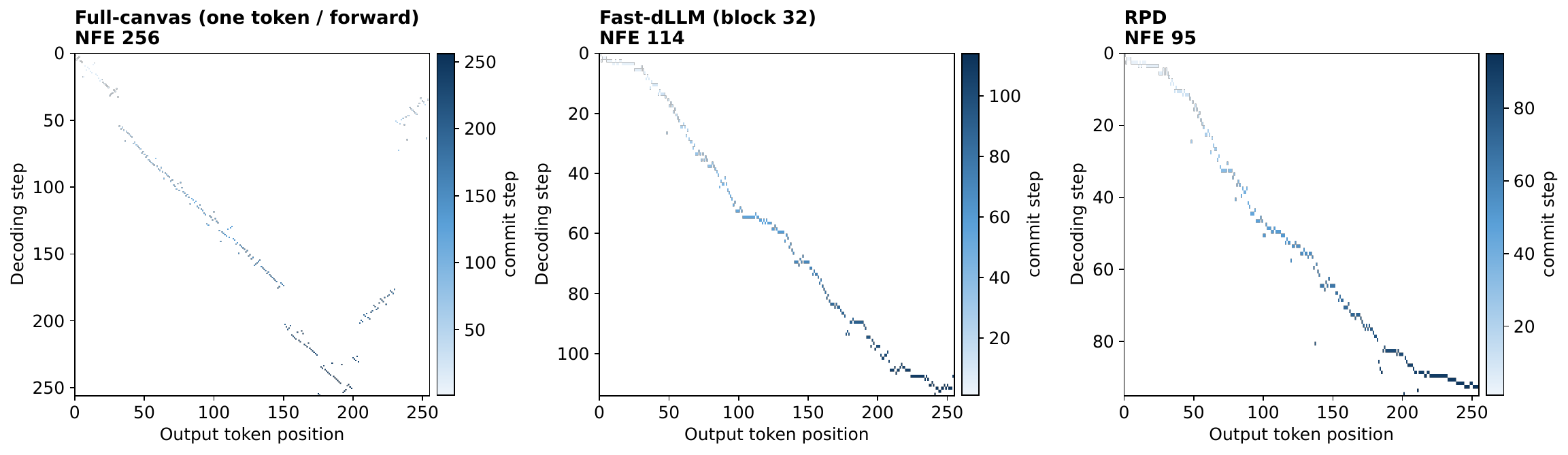}
\caption{Commit step of each output position under three decoders, LLaDA on GSM8K (first document). Lower is earlier.}\label{fig:traj_1x3}\end{figure}

Figure~\ref{fig:traj_1x3} places the three decoders side by side on the same
GSM8K document, plotting the step at which each output position is committed.
Committing one token per forward pass over the full canvas produces a broadly
left-to-right order, but the second half of the canvas receives scattered late
commits.  Fast-dLLM's block structure removes that scatter and yields a strict
block-by-block staircase, at the cost of never committing outside the active
block.  RPD recovers the same left-to-right progression on the full canvas
without any block schedule, and does so in fewer steps: $95$ forward passes
against $114$ for Fast-dLLM and $256$ for one-token-per-forward decoding.  The
cumulative entropy budget is therefore doing the work that the block schedule
does for Fast-dLLM, while remaining free to commit anywhere on the canvas when
the upstream context allows it.

\end{document}